\documentclass[runningheads]{llncs}
\usepackage{url}
\usepackage{graphicx}
\usepackage{amssymb}
\usepackage{latexsym}
\usepackage{multirow}
\usepackage{booktabs}
\usepackage{graphicx}
\usepackage{overpic}
\usepackage{bbm}
\usepackage{xcolor}
\usepackage{subfigure}
\usepackage[misc]{ifsym} 
\usepackage{graphicx}
\usepackage{mathrsfs}
\usepackage{multirow}
\usepackage{booktabs}
\usepackage{diagbox}
\usepackage{color}
\usepackage[colorlinks,linkcolor=blue,anchorcolor=blue,citecolor=blue,urlcolor=blue]{hyperref}
\usepackage[T1]{fontenc}

\usepackage{graphicx}
\usepackage{lipsum}
\newcommand\blfootnote[1]{%
  \begingroup
  \renewcommand\thefootnote{}\footnote{#1}%
  \addtocounter{footnote}{-1}%
  \endgroup
}

\begin{document}

\title{Gaze-directed Vision GNN for Mitigating Shortcut Learning in Medical Image}

\titlerunning{Gaze-directed Vision GNN}

\author{Shaoxuan Wu $^{\ddag,}$ \inst{1} \and Xiao Zhang $^{\ddag,*,}$ \inst{1} \and Bin Wang\inst{2} \and Zhuo Jin\inst{1} \and Hansheng Li\inst{1} \and Jun Feng $^{*,}$\inst{1}}
% index{Wu, Shaoxuan}
% index{Zhang, Xiao}
% index{Wang, Bin}
% index{Jin, Zhuo}
% index{Li, Hansheng}
% index{Feng, Jun}

\authorrunning{S. Wu et al.}


\institute{School of Information Science and Technology, Northwest University, Xi'an, China\\
\and
Northwestern University, Chicago IL 60611, USA\\
\email{xiaozhang@nwu.edu.cn, fengjun@nwu.edu.cn}\\
}

\maketitle              % typeset the header of the contribution

\begin{abstract}
Deep neural networks have demonstrated remarkable performance in medical image analysis. 
However, its susceptibility to spurious correlations due to shortcut learning raises concerns about network interpretability and reliability.
Furthermore, shortcut learning is exacerbated in medical contexts where disease indicators are often subtle and sparse.
In this paper, we propose a novel gaze-directed Vision GNN (called GD-ViG) to leverage the visual patterns of radiologists from gaze as expert knowledge, directing the network toward disease-relevant regions, and thereby mitigating shortcut learning.
GD-ViG consists of a gaze map generator (GMG) and a gaze-directed classifier (GDC). 
Combining the global modelling ability of GNNs with the locality of CNNs, GMG generates the gaze map based on radiologists' visual patterns.
Notably, it eliminates the need for real gaze data during inference, enhancing the network's practical applicability.
Utilizing gaze as the expert knowledge, the GDC directs the construction of graph structures by incorporating both feature distances and gaze distances, enabling the network to focus on disease-relevant foregrounds. Thereby avoiding shortcut learning and improving the network's interpretability.
The experiments on two public medical image datasets demonstrate that GD-ViG outperforms the state-of-the-art methods, and effectively mitigates shortcut learning. 
Our code is available at \href{https://github.com/SX-SS/GD-ViG}{https://github.com/SX-SS/GD-ViG}.

% \keywords{First keyword  \and Second keyword \and Another keyword.}
% \end{abstract}
\end{abstract}

% \footnotetext{$^{\ddag}$ indicates co-first author, $^{*}$ indicates co-corresponding author.}
% \let\thefootnote\relax\footnotetext{{$^{\ddag}$ indicates co-first author, $^{*}$ indicates co-corresponding author.}

\section{Introduction}
In recent years, deep learning methods have been applied to various fields, including computer vision \cite{resnet}, robotics \cite{robotics_Intro}, and medical image analysis \cite{recent_Intro,zhang2023anatomy}. 
However, its reliability and interpretability have consistently been questioned.
Research investigations have illuminated a critical issue: networks often learn spurious correlations caused by shortcut learning \cite{shortcut}, which refers to the model prioritizing learning simple but task-irrelevant content from the data, affecting generalizability and dependability.
\blfootnote{$^{\ddag}$ indicates co-first author, $^{*}$ indicates co-corresponding author.}
The phenomenon is particularly acute in the medical domain where disease regions tend to occupy a minimal proportion and low contrast, making their precise capture more challenging. 
For instance, in chest X-ray diagnosis, the pneumothorax region may account for only 1.36\% of the image, while the majority of image regions exhibit structural similarity \cite{pneumothorax_Intro,failures_Intro}. 
Networks may resort to suboptimal shortcuts like hospital-specific tokens \cite{shortcut}, thereby undermining the network's interpretability and reliability.

Several studies tackle the problem of shortcut learning by incorporating prior knowledge about the disease region. Such knowledge can guide the network to locate the abnormality accurately and reduce the reliance on spurious features. A common way to achieve this is to employ extra fine-grained annotations of abnormality, such as target bounding boxes \cite{Learning_Intro} or target masks \cite{tell_Intro}. However, it is essential to recognize that this strategy is resource-intensive and time-consuming.

Prior knowledge about the disease region can also be derived from radiologists' gaze, which indicates their visual cognitive behavior during diagnosis \cite{Exploring_Intro,Eye_Intro}. 
The eye gaze data obtained by eye trackers indicate the specific regions of interest for radiologists, which are also related to the disease and contain task-relevant knowledge \cite{Observational_Intro,Radiotransformer_Intro}. 
Moreover, they can be collected passively and cheaply during image reading, without incurring extra costs. 
Several methods have leveraged gaze to enhance the performance and interpretability of the network in medical image analysis \cite{GANet,EGViT,GazeGNN,Radiotransformer_Intro,MammoNet_Intro}. 
For instance, GA-Net \cite{GANet} employs the gaze map as supervision, encouraging the network to attend to the regions that humans focus on in medical images. 
EG-ViT \cite{EGViT} utilizes gaze to mask the harmful background shortcuts in medical images, rectifying shortcut learning. However, these methods only guide training, not testing, so the network may still be prone to shortcut learning when encountering unseen images. 
TSEN \cite{TSEN} and M-SEN \cite{MSEN} utilize GAN or biCLSTM for gaze generation and detection. 
However, they are constrained by the locality of CNNs, making it challenging to simultaneously consider lesion areas at different spatial positions.
By incorporating gaze embedding as network input, GazeGNN \cite{GazeGNN} improves the robustness of the network in inference, which also demonstrates that the interpretability of medical image analysis tasks can be improved by building graph nodes from image patches. 
However, a limitation of GazeGNN is it requires real gaze data during testing, which constrains the network's applicability and usability.

In this paper, we propose a novel gaze-directed Vision GNN (GD-ViG), an end-to-end method that integrates radiologists' gaze into the neural network. 
GD-ViG comprises two subnets: gaze map generator (GMG) and gaze-directed classifier (GDC). 
GMG learns the visual pattern of radiologists' reading by combining the global modeling ability of GNN and the locality of CNN to generate gaze maps. The need for real gaze in the inference stage is eliminated by GMG, enhancing the network's applicability.
GDC constructs the graph by fusing gaze distance and feature distance, eliminating connections to disease-irrelevant nodes from the graph structure. The proposed method is focused on task-relevant foreground by GDC, reducing shortcut learning, and improving its interpretability and robustness. 
Evaluations conducted on two public datasets demonstrate that GD-ViG outperforms the state-of-the-art methods.

\section{Method}
\begin{figure}[t]
\centering
\includegraphics[width=0.99\textwidth]{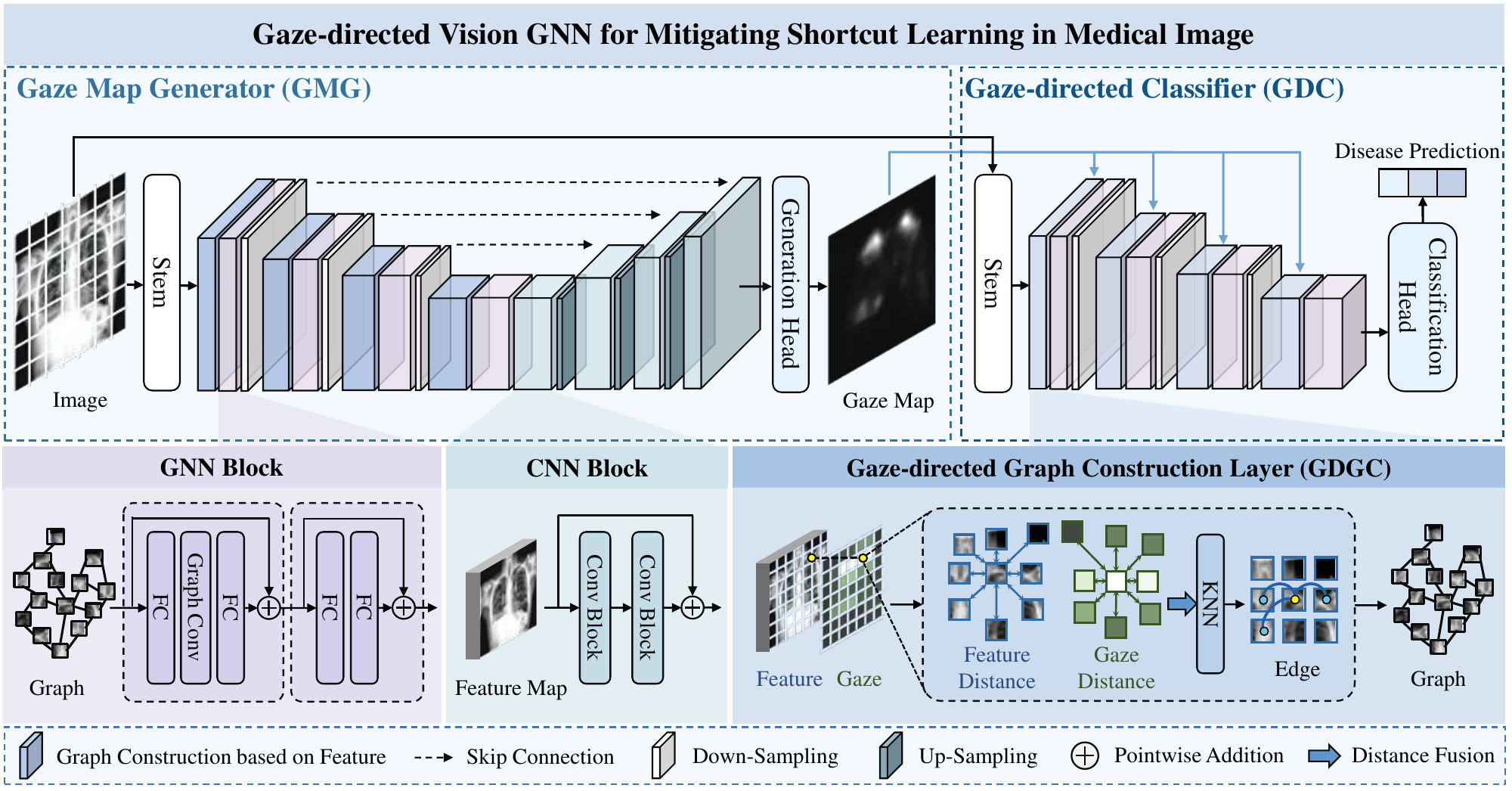}
\caption{Overview of the proposed framework consisting of two subnets: 1) Gaze map generator; 2) Gaze-directed classifier.} \label{fig1}
\end{figure}

\subsection{Framework for end-to-end GD-ViG} 
As illustrated in Fig.\,\ref{fig1}, the proposed GD-ViG is capable of assimilating the visual patterns of radiologists into the ViG, which consists of GMG and GDC. 
The GMG models the visual patterns of radiologists to generate the gaze map from the image.
Subsequently, the GDC leverages the generated gaze map to construct a graph structure, concentrating on disease-relevant regions for diagnosis. A notable feature of GD-ViG is its ability to generate the gaze map from the image during inference, thereby obviating the need for extra gaze data.

\subsection{Gaze map generator}
The real gaze map can be generated from raw gaze data by extracting fixation points and applying Gaussian blurring \cite{EGViT}.
Through supervised learning with real gaze maps, the GMG learns the visual pattern of radiologists and generates gaze maps from images.
The GMG comprises an encoder and a decoder. The encoder consists of graph construction layers based on feature distance and GNN blocks, which extract the global features of the image.
An $H\times W$ image is downsampled to a size of $\frac{H}{4} \times \frac{H}{4} \times C$ by the stem layer, where $C$ is the feature dimension. 
The graph construction layer then converts the feature at each position into a node, forming a node set 
$ {\mathcal V} = \{ {x_1}^C,{x_2}^C \cdots , {x_{N} }^C \}$, where $N$ is the number of nodes. 
For each node, the KNN algorithm searches its neighbors $ {\mathcal N}(x)$ based on the feature distance and connects them with an edge $e$. The edge set is denoted as  ${\mathcal E}$. Thus, a graph structure
$ {\mathcal G} = \{ {\mathcal N}, {\mathcal E}\}$ is constructed. 

The GNN blocks perform two operations: neighbor node feature fusion and feature transformation. The graph can be represented as a feature vector $X\in\mathbb{R}^{N\times C}$, and the neighbor node feature fusion operation can be formulated as:
\begin{equation}
    X^{'} = W_2 (  GC(  W_1 X)) + X,
\end{equation}
where $W_1(\cdot )$ and $W_2(\cdot )$ are fully connected layers and $GC(\cdot )$ is max-relative graph convolution \cite{MRGC}. The feature transformation can be formulated as:
\begin{equation}
    Y = W_4(W_3(X^{'}) + X^{'},
\end{equation}
where $W_3(\cdot )$ and $W_4(\cdot )$ are the fully connected layers. The decoder consists of four CNN blocks, which extract the local feature. The CNN block is as follows:
\begin{equation}
    Z = Conv_2(Conv_1(Y)) + Y,
\end{equation}
where $Conv_1(\cdot )$ and $Conv_2(\cdot )$ includes $3\times3$ convolution layer, batch normalization and ReLU. Skip connections are introduced in different blocks to fuse the global and local information. 

After the encoder and decoder, the image passes through the generation head, which outputs the gaze map ${gm}$ of size $H\times W$. 
The mean squared error loss is used as the GMG loss, which can be formulated as:
\begin{equation}
    {\mathcal L_{GMG}} = \frac{1}{H W}  {\textstyle \sum_{i=1}^{H\times W}} (gm_i - \hat{gm}_i) ^2,
\end{equation}
where $\hat{gm}$ is the ground truth gaze maps. 
The visual pattern of radiologists' reading is learned by GMG under the supervision of $\hat{gm}$. 
During inference, GMG is responsible for generating the gaze map to avoid the need for the real gaze.

\subsection{Gaze-Directed Classifier based on Gaze Distance}
As illustrated in Fig.\,\ref{fig1}, GDC comprises gaze-directed graph construction layers (GDGC) and GNN blocks. 
GDGC fuses feature distance and gaze distance to construct a graph structure and eliminate connections to disease-irrelevant nodes. 
It splits a feature map of $H_f\times W_f\times C$ into $H_f W_f$ nodes, denoted as 
$ {\mathcal V} = \{ {x_1}^C,{x_2}^C \cdots , {x_{N} }^C \}$ , where $x_i \in \mathbb{R}^{C}$ and $C$ is the feature dimension. 
The gaze map is also downsampled to the size of $H_f W_f$. 
The distance between central node $x_i$ and other nodes $x_j$ is defined as:
\begin{equation}
    dist(x_i, x_j) = {\parallel x_i - x_j\parallel}^2 + \lambda_g {\parallel gm_i - gm_j\parallel}^2 * gm_i,
    \label{eq_dist}
\end{equation}
where $\lambda_g$ is a hyperparameter. The $K$ neighbors of $x_i$, ${\mathcal N}(x_i)$, can be obtained using KNN. An edge is formed for each node and its neighbor, forming the edge set ${\mathcal E}$. The graph structure is then represented as $ {\mathcal G} = \{ {\mathcal N}, {\mathcal E}\}$. 
Fig.\, S1 of the supplementary material provides more details.
The gaze distance plays a larger role where $gm$ is large, making the nodes in the gaze highlighted area more connected, and then the GNN block can aggregate the information of these areas better. Feature distance and gaze distance serve as complementary information to construct a robust graph structure.
Fig.\,\ref{fig3} shows that following distance fusion, the connections with disease-irrelevant regions are eliminated, enhancing the network's interpretability and robustness.
The loss of GDC is the cross-entropy loss. It can be formulated as:
\begin{equation}
    % {\mathcal L_{GDC}} = \frac{1}{c}  {\textstyle \sum_{i=1}^{c}} \hat{y_i}~log y_i,
    {\mathcal L_{GDC}} = -{\textstyle \sum_{i=1}^{c}} \;\hat{y_i}\;log\;y_i,
\end{equation}
where $\hat{y}$ and $y$ are the ground truth and prediction of GDC and $c$ represents the number of categories. % \subsection{Loss Function} 
The loss function of GD-ViG is composed of the loss of GMG and the loss of GDC, as follows: 
\begin{equation}
    {\mathcal L} = {\mathcal L_{GMG}} + \lambda_c {\mathcal L_{GDC}},
    \label{eq_loss}
\end{equation}
where $\lambda_c$ is the balance coefficient. 
% GD-ViG is trained in an end-to-end manner.

\section{Experiments and Results}
\subsection{Dataset and Evaluation Metrics}

The proposed GD-ViG was evaluated on two datasets: SIIM-ACR \cite{SIIMACR} and EGD-CXR \cite{EGDCXR}. SIIM-ACR contains 1170 chest X-ray images, of which 268 have pneumothorax and the corresponding gaze data. EGD-CXR consists of 1083 chest X-ray images, sourced from the MIMIC-CXR \cite{MIMIC} dataset. EGD-CXR annotates the chest images into three categories: Normal, Congestive Heart Failure, and Pneumonia, and the gaze data are provided for each image. 
Each image was resized to 224×224 and a two-dimensional Gaussian smoothing was applied to transform the gaze into the gaze map, following \cite{EGViT}.

Accuracy (ACC), area under the receiver operating characteristic curve (AUC), and F1 score (F1) were used as metrics to evaluate the performance. 
All experiments were implemented using Pytorch on a single NVIDIA 3080Ti GPU (12GB). 
The network was trained using Adam optimizer with an initial learning rate of $10^{-4}$ for $100$ epochs. We set $\lambda_g=3$ in Eq. \ref{eq_dist} and $\lambda_c=1$ in Eq. \ref{eq_loss}.

\subsection{Comparison with State-of-the-art Methods}

We performed quantitative and qualitative evaluations of GD-ViG and compared it with various state-of-the-art methods. 
The methods were categorized into three groups: 1) Methods without gaze: ResNet \cite{resnet}, Vision Transformer \cite{ViT}, Swin Transformer \cite{swinT}, and Vision GNN \cite{ViG}; 2) Methods with gaze during training: M-SEN \cite{MSEN}, EML-Net \cite{EMLNet}, ResNet+Gaze \cite{GANet} and EG-ViT \cite{EGViT}; 
3) Methods with gaze during inference: GazeGNN \cite{GazeGNN}.

\begin{table}[ht]
%\scriptsize
   \centering%\multicolumn{3}{ r}
       \caption{Comparison with other methods. Bold indicates the best result.}
   \resizebox{115mm}{35mm}{
   \setlength{\tabcolsep}{1.5mm}{
   \begin{tabular}{c c c c c c c}
    \hline
    \multirow{2}{*}{\textbf{Method}}
    & \multicolumn{3}{c}{\textbf{SIIM-ACR}}  
    & \multicolumn{3}{c}{\textbf{EGD-CXR}}
    \\
    \cmidrule(r) {2 - 4}
    \cmidrule(r) {5 - 7}
    & \textbf{Acc$\uparrow$} & \textbf{AUC$\uparrow$} 
    & \textbf{F1$\uparrow$}  & \textbf{Acc$\uparrow$} 
    & \textbf{AUC$\uparrow$} 
    & \textbf{F1$\uparrow$} \\
    % \hline
    \hline
    
     ResNet-18 \cite{resnet}
     & 83.20  & 82.35  & 85.25 
     & 71.96  & 85.02  & 72.17 \\
     ResNet-50 \cite{resnet}
     & 84.00  & 85.81  & 83.63 
     & 72.90  & 86.43  & 72.45 \\
     ResNet-101 \cite{resnet}
     & 84.40  & 86.13  & 81.08 
     & 74.77  & 85.63  & 74.90  \\
     ViT \cite{ViT}
     & 83.60  & 84.16  & 83.77 
     & 70.09  & 85.43  & 69.19  \\
     SwinT \cite{swinT}
     & 84.40  & 83.31  & 83.69 
     & 71.96  & 86.44  & 74.89  \\
     ViG \cite{ViG}
     & 83.20  & 84.89  & 82.81 
     & 75.70  & 85.71  & 75.62 \\
     \hline 

     M-SEN \cite{MSEN}
     & 84.80  & 85.93  & 84.03  
     & 78.50  & 84.27  & 77.45 \\
     EML-Net \cite{EMLNet}
     & 85.20  & 83.65  & 85.25  
     & 77.57  & 87.33  & 75.47 \\
     ResNet-18+Gaze \cite{GANet}
     & 84.80  & 71.26  & 83.71  
     & 77.57  & 86.13  & 77.47 \\
     ResNet-50+Gaze \cite{GANet}
     & 83.20  & 70.25  & 82.35  
     & 78.50  & 86.43  & 77.91 \\
     ResNet-101+Gaze \cite{GANet}
     & 84.80  & 72.68  & 84.03  
     & 79.44  & 86.42  & 79.20 \\
     EG-ViT \cite{EGViT}
     & 85.60  & 75.30  & 85.14  
     & 77.57  & 85.53  & 77.42\\
     \textbf{Ours} 
     & \textbf{87.20}  & \textbf{86.99}  & \textbf{86.68}  
     & \textbf{85.05}  & 88.56  & \textbf{84.53}  \\
     \hline
     GazeGNN \cite{GazeGNN}
     & 85.60  & 85.16  & 85.60  
     & 83.18  & \textbf{92.30}   & 82.30  \\
     \hline
	\end{tabular}}
    }
    \label{tab1}
\end{table}

\noindent{\textbf{Quantitative Results.} 
The quantitative results are presented in Table.\ref{tab1} demonstrate that our method achieves SOTA performance on the SIIM-ACR dataset, achieving an accuracy of 87.2\%. This represents a 1.6\% improvement over the previously best-performing GazeGNN method. 
Furthermore, the highest accuracy on the EGD-CXR dataset is also achieved by the proposed method, recorded at 85.05\%. 
In contrast to GazeGNN, which necessitates real gaze data during inference, our method is capable of generating a gaze map via GMG. The ability of GMG to produce gaze maps comparable to real gaze, coupled with GDC's effective utilization of gaze as prior information, is thereby indicated.
Moreover, the method and other methods have $p$-values less than 0.05 in paired $t$-tests on Acc and F1 metrics.

\begin{figure}[!t]
\centering
\includegraphics[width=0.975\textwidth]{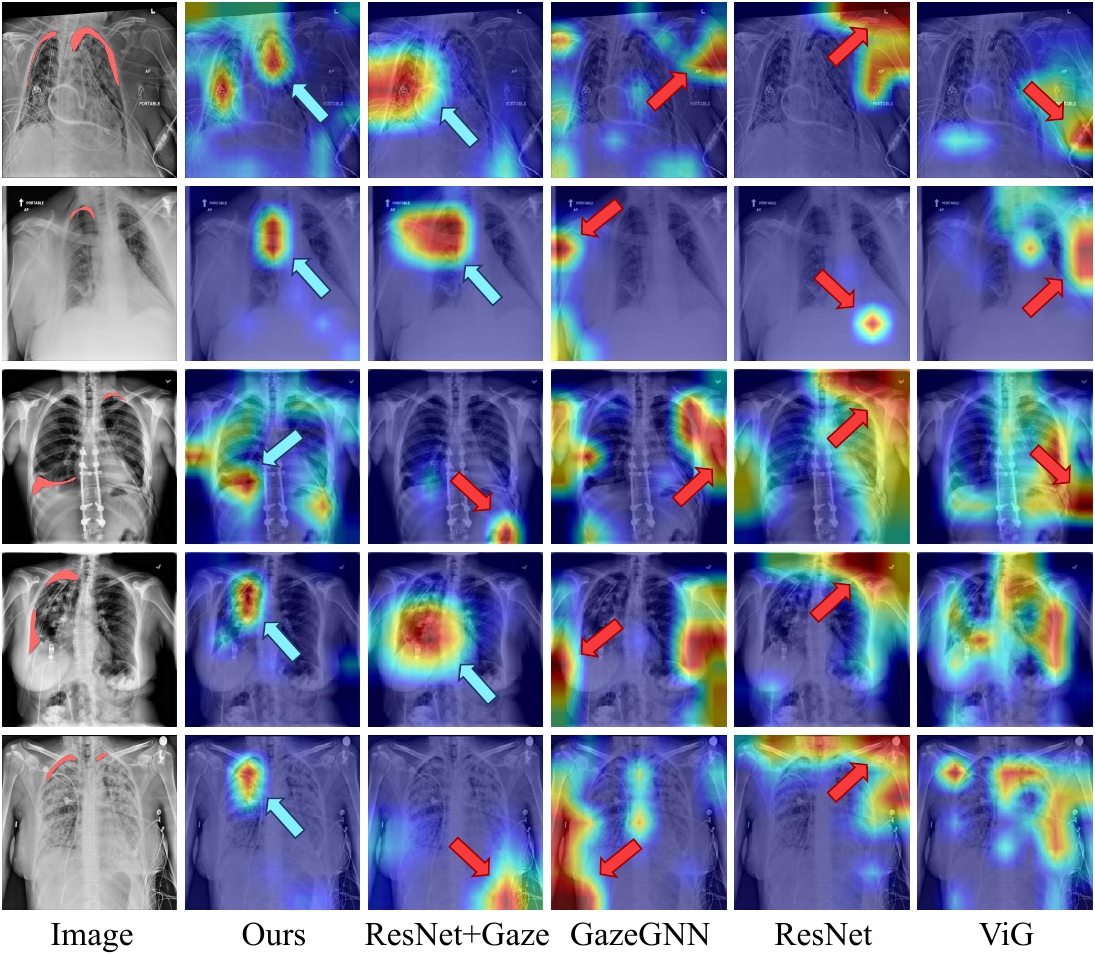}
\caption{Comparison results of attention map of different methods on SIIM-ACR dataset. In the first column, the pneumothorax region is marked in red mask. Networks prone to shortcut learning are indicated by red arrows, while blue arrows denote regions where networks accurately concentrate on disease-specific areas.} 
\label{fig2}
\end{figure}

\begin{table}[ht]
   \centering
   \caption{Quantitative results of ablation analysis of different components.}
   % \vspace{-1mm}
   \resizebox{110mm}{14mm}{
   \setlength{\tabcolsep}{1.5mm}{
   \begin{tabular}{c c c c c c c}
    \hline
    \multirow{2}{*}{\textbf{Method}}
    & \multicolumn{3}{c}{\textbf{SIIM-ACR}}  
    & \multicolumn{3}{c}{\textbf{EGD-CXR}}
    \\
    \cmidrule(r) {2 - 4}
    \cmidrule(r) {5 - 7}
    & \textbf{Acc$\uparrow$} & \textbf{AUC$\uparrow$} 
    & \textbf{F1$\uparrow$}  & \textbf{Acc$\uparrow$} 
    & \textbf{AUC$\uparrow$} 
    & \textbf{F1$\uparrow$} \\
    \hline
    \
     ViG
     & 83.20  &  84.89   &  82.81 
     & 75.70  &  85.71   &  75.62 \\
     ViG+AMG(CNN)
     & 85.60  & ~\textbf{86.99 }   &  85.14   
     & 80.37  & 86.07  &  80.39 \\
     ViG+AMG(GNN) 
     &  86.00   & 86.35   & 85.62
     &  83.18   & 86.52   & 83.15   \\
     ViG+AMG(GNN+CNN) 
     & \textbf{87.20}  & \textbf{86.99}  & \textbf{86.68}  
     & \textbf{85.05}  & \textbf{88.56}    
     & \textbf{84.53}  \\
     \hline
	\end{tabular}}
    }
    \label{tab2}
\end{table}

\noindent{\textbf{Qualitative Visualization.}
Grad-CAM \cite{GradCAM} is utilized to visualize the attention maps of various methods on the SIIM-ACR dataset, as shown in Fig.\,\ref{fig2}. 
Attention maps on the EGD-CXR data are similarly visualized in Fig.\,S3 of the supplementary material. 
The regions of focus for the networks are indicated by red areas, with the first column displaying the original image and the pneumothorax disease area highlighted in red mask.
A comparison of the results from different methods on the SIIM-ACR data reveals that networks lacking gaze direction concentrate on regions not pertinent to the disease, as indicated by red arrows. Conversely, even with gaze direction, GazeGNN and ResNet+Gaze exhibit tendencies of shortcut learning. Our method mitigates this issue and accurately targets the disease area, marked by a blue arrow. Additionally, Fig.\,S2 of the supplementary material presents a comparison of the gaze generated by the proposed method against other methods, demonstrating the ability of the proposed method to produce a high-quality and realistic gaze map.

\subsection{Ablation Study}
Ablation experiments were conducted on the SIIM-ACR and EGD-CXR datasets, with results reported in Table.\ref{tab2}. The impact of gaze data utilization and GMG structure was investigated. Without any strategy, the ViG accuracy stood at 83.20\%. The incorporation of gaze, generated by CNN, elevated accuracy to 85.60\%, signifying GDC's capacity to effectively harness gaze information for performance enhancement. Accuracy further climbed to 86.00\% when gaze maps were produced by GNN. Utilizing both GNN and CNN for gaze map generation culminated in the highest accuracy of 87.20\%, surpassing all other methods. Paired $t$-tests were executed to evaluate the performance enhancement attributable to the proposed modules, including Acc, AUC, and F1, with all the $p$-values lower than 0.05. These results affirm the GMG's proficiency in amalgamating global and local information to refine gaze map generation.

\begin{figure}[!t]
\centering
\includegraphics[width=0.975\textwidth]{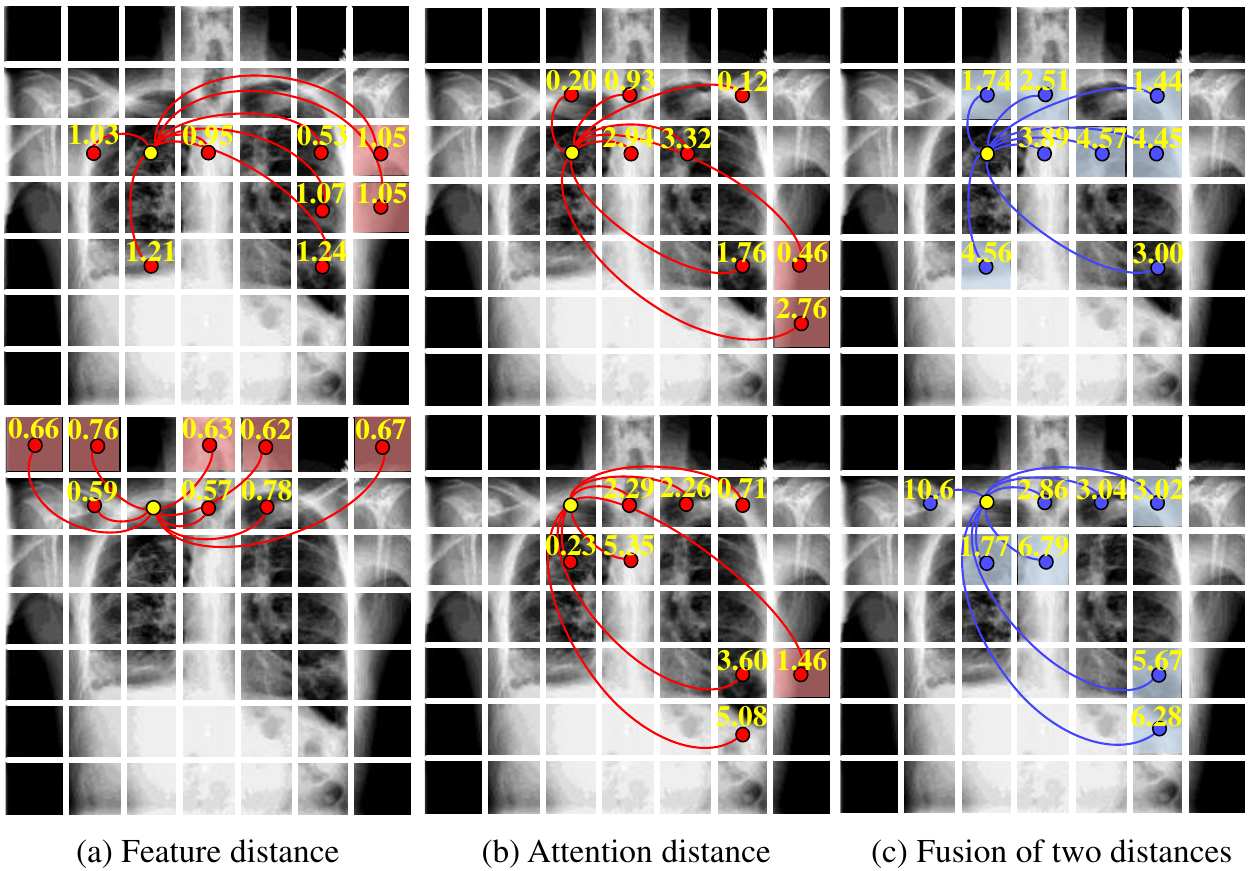}
% \vspace{-1mm}
\caption{
Visualization of the graph structure based on different distances. The red regions and the blue regions represent the nodes that are eliminated and the disease-relevant nodes that are retained after distance fusion, respectively.
} \label{fig3}
\end{figure}

\subsection{Graph Structure Visualization}
% \vspace{-2mm}
Fig.\,\ref{fig3} illustrates the synergistic effect of gaze and feature distances in graph construction. 
Displayed across three columns, each based on distinct distances, the yellow circles denote central nodes with their neighboring nodes connected. Quantitative values of the distance between nodes are marked with yellow numbers.
Central nodes reliant on feature and gaze distances are observed to connect to disease-irrelevant areas, highlighted in red. After the fusion of distances, these connections are severed. Concurrently, disease-relevant nodes, accentuated in blue within the initial graphs, are preserved, underscoring the complementary nature of the two distances. The distance fusion strategy introduced herein facilitates the formation of a graph structure enriched with interconnected foreground nodes, diminishing the impact of extraneous regions, curtailing shortcut learning, and enhancing interpretability.

\section{Conclusion}

In this paper, we propose GD-ViG, an end-to-end neural network that integrates the radiologists' gaze as prior information into Vision GNN to mitigate shortcut learning.
GD-ViG consists of two subnets: gaze map generator and gaze-directed classifier. 
The gaze map generator models radiologists' visual patterns through the global modeling capabilities of GNN and CNN's local feature extraction, thereby generating the gaze map. 
It obviates the necessity for the real gaze data during inference, broadening the applicability of the network. 
The gaze-directed classifier employs the gaze map as expert prior knowledge for graph construction, severing connections with nodes unrelated to the disease and concentrating on pertinent regions. 
Such a method not only augments the network's interpretability but also precludes shortcut learning. 
Evaluations conducted on the SIIM-ACR and EGD-CXR datasets demonstrate that our method outperforms the state-of-the-art methods and significantly enhances the network's interpretability and reliability.

\begin{credits}
\subsubsection{\ackname} 
This work is supported by the National Natural Science Foundation of China (NSFC Grant No. 62073260).

\subsubsection{\discintname}
The authors have no competing interests to declare that are relevant to the content of this article.

\end{credits}

\end{document}